\documentclass[11pt, a4paper, logo, two column]{versions/craftjarvis}

\pdftrailerid{redacted}

\usepackage[authoryear, sort&compress, round]{natbib} 
\usepackage[pagebackref,breaklinks,colorlinks,allcolors=iccvblue]{hyperref}

\renewcommand{\today}{} 

\newcommand{\arxivnote}{The project page is available at \url{https://craftjarvis.github.io/ROCKET-2/}.}

\definecolor{iccvblue}{rgb}{0.21,0.49,0.74}

\usepackage{graphicx}
\usepackage{times}
\usepackage{url}
\usepackage{natbib}
\usepackage{caption}
\usepackage{relsize}

\usepackage{amssymb}
\usepackage{amsmath}

\usepackage{tabularray}
\UseTblrLibrary{booktabs,varwidth}
\usepackage{makecell}

\usepackage{ragged2e}
\usepackage{enumitem}
\usepackage{cleveref}
\usepackage[export]{adjustbox}
\usepackage{xspace}

\usepackage{cuted}
\usepackage{arydshln}
\usepackage{marvosym}
\usepackage{xcolor}

\usepackage{common/notation}

\usepackage{algorithm}
\usepackage{algorithmic}

\usepackage{newfloat}
\usepackage{listings}

\definecolor{Gray}{gray}{0.9}
\definecolor{mygreen}{rgb}{0.0, 0.5, 0.0}
\definecolor{myred}{rgb}{0.8, 0.25, 0.33}
\definecolor{myblue}{rgb}{0.19, 0.55, 0.91}
\definecolor{uclablue}{rgb}{0.15, 0.45, 0.68}
\definecolor{boxgreen}{rgb}{0.02, 0.66, 0.02}
\definecolor{boxred}{rgb}{0.66, 0.1, 0.1}
\definecolor{boxblue}{rgb}{0.01, 0.01, 0.73}
\definecolor{mygray}{gray}{0.4}

\usepackage{amsmath,amsfonts,bm}

\def\eqref#1{equation~\ref{#1}}

\def\1{\bm{1}}

\DeclareMathAlphabet{\mathsfit}{\encodingdefault}{\sfdefault}{m}{sl}
\SetMathAlphabet{\mathsfit}{bold}{\encodingdefault}{\sfdefault}{bx}{n}

\newcommand{\papertitle}{\agent: Steering Visuomotor Policy via Cross-View Goal Alignment}

\newcommand{\agent}{\textsc{ROCKET-2}\xspace}

\newif\ifaaai
\newif\ifarxiv
\newif\ificcv

\newcommand{\aaaionly}[1]{\ifaaai#1\fi}
\newcommand{\arxivonly}[1]{\ifarxiv#1\fi}
\newcommand{\iccvonly}[1]{\ificcv#1\fi}

\newcommand{\ifversion}[2]{%
  \expandafter\ifx\csname if#1\endcsname\iftrue
    #2%
  \fi
}

\renewcommand{\citestyle}{%
  \aaaionly{\bibliographystyle{aaai2026}}%
  \arxivonly{\bibliographystyle{plainnat}}%
  \iccvonly{\bibliographystyle{ieee_fullname}}%
}

\graphicspath{{figures/}{assets/}{tables/}}

\usepackage{xcolor}
\definecolor{Gray}{gray}{0.9}
\definecolor{mygreen}{rgb}{0.0, 0.5, 0.0}
\definecolor{myred}{rgb}{0.8, 0.25, 0.33}
\definecolor{myblue}{rgb}{0.19, 0.55, 0.91}
\definecolor{uclablue}{rgb}{0.15, 0.45, 0.68}
\definecolor{boxgreen}{rgb}{0.02, 0.66, 0.02}
\definecolor{boxred}{rgb}{0.66, 0.1, 0.1}
\definecolor{boxblue}{rgb}{0.01, 0.01, 0.73}
\definecolor{mygray}{gray}{0.4}

\usepackage{notation}
\usepackage{times}
\usepackage{cuted}
\usepackage{arydshln}
\usepackage{marvosym}
\usepackage{graphicx}
\usepackage{amssymb}
\usepackage{url}

\usepackage{microtype}
\usepackage{subcaption}
\usepackage{booktabs}

\usepackage{amsmath}
\usepackage{mathtools}
\usepackage{amsthm}

\usepackage{listings}
\usepackage{etoc}
\usepackage{algorithm}
\usepackage{algorithmic}
\usepackage{lipsum}
\usepackage{pifont}
\usepackage{colortbl}
\usepackage{acronym}
\usepackage{multirow}
\usepackage{makecell}
\usepackage{enumitem}
\usepackage{tabularx}
\usepackage{colortbl}
\usepackage{array}
\usepackage[skip=3pt,font=small]{caption}
\definecolor{iccvblue}{rgb}{0.21,0.49,0.74}
\usepackage{xspace}
\usepackage{mdframed}

\usepackage[export]{adjustbox}

\newcolumntype{Y}{>{\arraybackslash}X}

\usepackage[utf8]{inputenc} 
\usepackage[T1]{fontenc}    
\usepackage{amsfonts}       
\usepackage{nicefrac}       

\usepackage{mathtools}
\usepackage{amssymb}
\usepackage{amsthm}

\renewcommand{\paragraph}[1]{\noindent\textbf{#1.}}

\renewcommand{\paragraph}[1]{\noindent\textbf{#1.}}

\makeatletter
\DeclareRobustCommand\onedot{\futurelet\@let@token\@onedot}
\def\@onedot{\ifx\@let@token.\else.\null\fi\xspace}

\def\wrt{w.r.t\onedot}

\makeatother

\acrodef{llms}[LLMs]{Large Language Models}
\acrodef{mlms}[MLMs]{Multimodal Language Models}

\usepackage{xcolor}

\arxivtrue

\title{\papertitle}

\author[1]{Shaofei~Cai}
\author[1]{Zhancun~Mu}
\author[2]{Anji~Liu}
\author[1]{Yitao~Liang \textrm{\Letter}}

\affil[1]{Peking~University}
\affil[2]{University~of~California,~Los Angeles}
\affil[ \hspace{-0.73ex}]{All authors are affiliated with Team CraftJarvis} 

\begin{document}

\begin{abstract}
We aim to develop a goal specification method that is semantically clear, spatially sensitive, domain-agnostic, and intuitive for human users to guide agent interactions in 3D environments. Specifically, we propose a novel cross-view goal alignment framework that allows users to specify target objects using segmentation masks from their camera views rather than the agent’s observations. We highlight that behavior cloning alone fails to align the agent’s behavior with human intent when the human and agent camera views differ significantly. To address this, we introduce two auxiliary objectives: cross-view consistency loss and target visibility loss, which explicitly enhance the agent's spatial reasoning ability. 
According to this, we develop \agent, a state-of-the-art agent trained in Minecraft, achieving an improvement in the efficiency of inference $3\times$ to $6\times$ compared to ROCKET-1. 
We show that \agent can directly interpret goals from human camera views, enabling better human-agent interaction. 
Remarkably, ROCKET-2 demonstrates zero-shot generalization capabilities: despite being trained exclusively on the Minecraft dataset, it can adapt and generalize to other 3D environments like Doom, DMLab, and Unreal through a simple action space mapping. 
\arxivnote
\end{abstract}

\twocolumn[{%
\renewcommand\twocolumn[1][]{#1}%
\maketitle
\vspace{-0.5cm}
\centering
\includegraphics[width=0.99\linewidth]{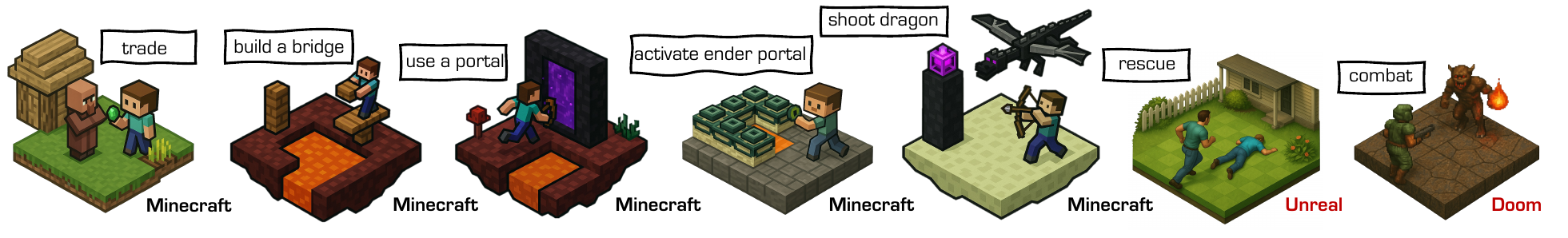}
\captionof{figure}{
\textbf{Powered by cross-view goal specification, we are the first to show that AI agents can complete complex tasks such as \textit{building a bridge} and \textit{damaging the dragon} in Minecraft. In addition, it demonstrates an impressive zero-shot generalization to other 3D games.} 
}
\vspace{1em}
\label{fig:teaser}
}]

\section{Introduction}
\label{sec:intro}
Learning an agent to achieve desired goals is a long-standing challenge in the field of embodied intelligence, with significant implications for the development of robots \citep{rt-1, rt-2, bcz} and virtual players \citep{deps, jarvis-1, voyager}. A key challenge is to find goal representations that are (i) flexible for human users to specify and (ii) expressive and precise to capture as many tasks as possible. Most current approaches address only one of these aspects. 
For example, traditional works \citep{rt-1, palm-e, language_table} focus on training agents to follow language instructions. As pointed out in \cite{rt-sketch, cai2024rocket}, while language is intuitive, it relies on numerous prepositions to express spatial relationships, which can be vague and inefficient. Furthermore, language also suffers from the generalization problem of novel visual concepts \citep{groot-1}. 
Realizing these limitations, some works attempted to introduce visual modalities into goal representations. For example, \cite{rt-sketch} employs hand-drawn target layouts in robot manipulation environments to represent human intent; \cite{rt-trajectory} uses end-effector trajectory sketches for fine-grained control of robot arms; and ROCKET-1~\citep{cai2024rocket} specifies the objects to interact with by applying segmentation masks to the agent’s perception. These methods greatly improved the expressiveness of spatial relationships and generalization across tasks. 
However, both trajectory sketches and object segmentation are closely tied to the agent’s current observation, causing issues in partially observable 3D worlds. These include: (i) goals need to be generated in real-time as the agent's camera view changes; (ii) goals cannot be specified when the target is occluded. 

To strike a balance between expressiveness and flexibility, we propose an innovative and user-friendly cross-view goal specification method. 
It allows human users to specify the target object using segmentation masks from their own camera view, rather than the agent’s camera view. The agent is then trained to align with human intent and take actions based on its own observations via imitation learning. Decoupling the goal specification from the camera view of the agent will significantly enhance the efficiency of human-agent interaction. 
However, the partial observability of open worlds makes aligning goals across camera views challenging. This involves handling occlusion, geometric deformation, and the distinction of objects of similar look. 
In Figure \ref{fig:teaser}, we show an agent in the left corner and a human player standing on a farmland. The human intends to command the agent to hunt a sheep near the house, even though the agent cannot initially observe the target sheep. To achieve this, the agent must establish spatial relationships using shared visual landmarks between the human’s camera view and its own. We find that relying solely on a behavior cloning loss is insufficient. 

To address these challenges, we highlight an important property of behavior datasets \citep{cai2024rocket}: \textbf{The target object remains consistent across camera views in an interaction trajectory}. Motivated by this, we propose two auxiliary objective functions: \textit{cross-view consistency loss} and \textit{target visibility loss}, to explicitly enhance the agent’s ability to align goals across camera views. Specifically, cross-view consistency loss requires the agent to accurately predict the target object’s centroid point \wrt its camera view, while target visibility loss helps the agent determine whether the target object is occluded. 
By combining these auxiliary losses with behavior cloning loss, we develop \agent, a state-of-the-art agent in Minecraft. 
Our experiments show that \agent can autonomously track the target object as the camera changes, eliminating the need for SAM’s \citep{sam2} real-time semantic segmentation, speeding up inference $3\times$ to $6\times$ compared to ROCKET-1. 
We observe that \agent can interpret intentions from a human's camera view and make decisions to achieve expected goals in the 3D world. 
Notably, by simply mapping Minecraft's action space to that of other 3D environments, such as Doom~\citep{vizdoom}, DMLab~\citep{dmlab}, and Unreal Engine~\citep{unrealzoo}, we are surprised to find that \textbf{\agent could successfully perform basic navigation and object interaction tasks in a zero-shot generalization manner. }
This demonstrates that our proposed cross-view alignment scheme is domain-agnostic and holds large potential for further scaling across a wider range of environments. 

\textbf{Our contributions are threefold}: \textbf{(1)} We introduce a user-friendly interface that allows humans to specify goals using instance-segmentation from humans' camera view. \textbf{(2)} By introducing \textit{cross-view consistency loss} and \textit{target visibility loss}, we train \agent, an agent that autonomously tracks targets, eliminating the need for real-time goal segmentation and significantly speeding up inference. \textbf{(3)} We demonstrate that \agent, trained solely on Minecraft data, can zero-shot generalize to other 3D environments, which opens new avenues for discovering universal decision representations in 3D environments. 


\section{Related Works}
\label{sec:related}

\paragraph{Partial Observability} 
We address policy learning in partially observable 3D environments \citep{habitat, MineStudio}, where the agent perceives only egocentric views and must actively explore to locate key objects \citep{csgo}. Some methods leverage 3D point clouds for global context \citep{leo, jiang2024bevnav}, but such data is often unavailable. More commonly, memory-based architectures are used to aggregate past observations: RNNs help avoid redundant exploration \citep{zhao2023zero, gadre2022clip}, and TransformerXL enables long-horizon planning in Minecraft \citep{vpt, minerl}. 
To mitigate partial observability, some works \citep{krantz2023navigating, krantz2022instance} use instance images to specify goals, but these require unoccluded, centered views where the object occupies most of the image—constraints that limit their applicability. Instead, we require the policy to align object references across egocentric views, enabling robust target tracking under camera motion. While cross-view alignment has been explored in computer vision for BEV segmentation \citep{borse2023x} and re-identification \citep{reid}, we are the first to apply it to open-world policy learning.

\paragraph{Goal-Conditioned Imitation Learning} 
GCIL optimizes conditional policies via behavior cloning \citep{bc}, using goal representations such as language \citep{rt-1, rt-2, language_table}, images \citep{zson, steve1, rt-sketch}, videos \citep{groot-1, groot-2}, or trajectory sketches \citep{mimicplay, rt-trajectory}. Compared to standard imitation learning, GCIL provides clearer targets, simplifies behavior modeling, and improves policy controllability. 
Language goals are commonly used \citep{rt-x, palm-e, jarvis-1} but often fail to convey spatial details \citep{cai2024rocket, rt-trajectory}. Image goals \citep{zson, mimicplay} better ground spatial intent but are sensitive to lighting, textures, and other irrelevant features. Sketches \citep{rt-sketch} reduce visual sensitivity but are difficult to generate. Trajectory sketches \citep{rt-trajectory} improve control and generalization, but assume full observability, limiting their use in 3D worlds. We instead propose aligning goal-relevant objects across views, enabling spatially grounded yet robust goal conditioning in partially observable environments.

\begin{figure*}[t]
\begin{center}
\includegraphics[width=0.99\linewidth]{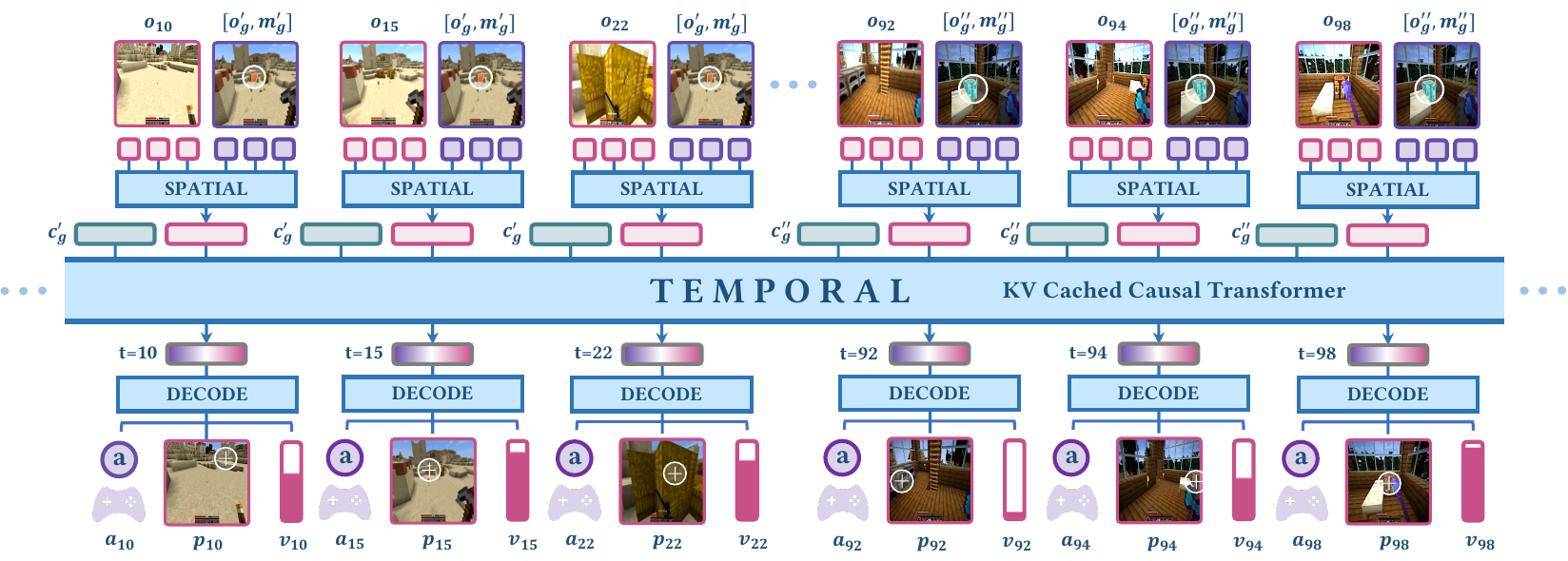}
\vspace{-2mm}
\end{center}
\caption{
\textbf{Overview of our approach.} We address the challenge of steering visuomotor policy via cross-view goal alignment. Our method allows humans to specify goals from their camera view while the agent acts based on its own observations.
}
\label{fig:pipeline}
\end{figure*}

\section{Method}
\label{sec:method}
In this section, we first introduce the problem of cross-view segmentation-conditioned policy, discussing it from the perspective of imitation learning. Next, we describe the process of generating cross-view trajectories annotated with semantic segmentation. We then present two auxiliary objectives designed to enhance cross-view object alignment in 3D scenes: the \textit{cross-view consistency loss} and the \textit{target visibility loss}. Finally, we detail the architecture of ROCKET-2 and outline the overall optimization objectives. 

\paragraph{Problem Statement} Our goal is to learn a goal-conditioned visuomotor policy, which allows humans to specify goal objects for interaction using semantic segmentation across camera views. Formally, we aim to learn a policy $\pi_{\text{cross}}(a_t | o_{1:t}, \{o_g, m_g\}, c_g)$, where $a_t$ represents the action at time $t$, $c_g$ denotes the interaction event, such as \textit{break item}, \textit{kill entity}, and \textit{approach}. In the Minecraft environment, an action corresponds to raw mouse and keyboard inputs. $o_t \in \mathbb{R}^{H \times W \times 3}$ denotes the environment observation at time $t$, and $o_g \in \mathbb{R}^{H \times W \times 3}$ represents an observation of the local environment from a specific camera view. Generally, $o_g$ and $o_t$ have some visual content overlap. $m_g \in \{0, 1\}^{H \times W \times 1}$ is a segmentation mask for $o_g$, highlighting the target object within the camera view $o_g$. 
During inference, users select a view $o_g$ containing the desired object from historical observations returned by the environment and generates its corresponding semantic segmentation $m_g$. To train such visuomotor policy, we assume access to a dataset $\mathcal{D}_{\text{cross}} = \{c^n, (o_t^n, a_t^n, m_t^n)_{t=1}^{L(n)}\}_{n=1}^{N}$ consisting of $N$ successful demonstration episodes, $L(n)$ is the length of episode $n$. \textbf{Within each episode, if $m_t$ is non-empty, all $(o_t, m_t)$ pairs indicate the same object.} Consequently, we can arbitrarily pick one observation frame as the goal view condition for the entire trajectory. 

\paragraph{Cross-View Dataset Generation} \label{dataset-generation} Without loss of generality, we use the Minecraft world as an example to illustrate the data generation process. Manually collecting datasets that meet the requirements is highly expensive. Thus, we employ the \textit{backward trajectory relabeling} technique proposed in \cite{cai2024rocket} to automate the annotation of the OpenAI Contractor Dataset \citep{vpt}, which consists of free-play trajectories from human players: $\mathcal{D}_{\text{raw}} = \{(o_t^n, a_t^n)_{t=1}^{L(n)}\}_{n=1}^{N}$. Specifically, for any given episode $n$, we first detect all frames $o^{n}_j$ where interaction events occur, identify the interaction type $c^{n}_j$, and localize the interacted object near frame $j$ using bounding boxes and point-based prompts. The SAM-2 \citep{sam2} model is then employed to generate the segmentation mask $m^{n}_j$ for the object. 
Starting from frame $j$, we traverse the trajectory backward and use the SAM-2 model to continuously generate segmentation masks for the object in real-time until either a new interaction event is encountered or a maximum tracking length is reached. Let $i$ denote the end frame. The resulting trajectory clip is then added to the training dataset: 
$
\mathcal{D}_{\text{cross}} \leftarrow \mathcal{D}_{\text{cross}} \cup \{c_j, (o^n_t, a^n_t, m^n_t)_{t=i}^{j}\}.
$
This ensures that every extracted clip is associated with a consistent interaction intent. The generated data encompasses the fundamental interaction types in Minecraft, including \textit{use}, \textit{break}, \textit{approach}, \textit{craft}, and \textit{kill entity}. Among these, \textit{approach} is a unique event, identified by detecting trajectory clips where the displacement exceeds a specified threshold. The object located at the center of the clip's final frame is designated as the goal of the \textit{approach} event. 

\paragraph{Cross-View Consistency Loss} Accurately interpreting the cross-view goal requires the policy to possess cross-view visual object alignment ability in 3D scenes. To achieve this, the model must fully exploit visual cues from different camera views, such as scene layout and landmark buildings, while being robust to challenges like occlusion, shape variations, and changes in distance. We observe that relying solely on behavior cloning loss \citep{bc} is insufficient. Therefore, we propose a \textit{cross-view consistency loss}. Since the segmentation across different camera views corresponds to the same object, we train the model to condition on the segmentation from one camera view to generate the segmentation for another camera view, thereby directly enhancing the model's 3D spatial perception. 
To reduce computational complexity, we opt to predict the centroid of the segmentation mask instead of the complete mask, formally expressed as:
$\pi_{\text{cross}}(p_t \mid o_{1:t}, \{ o_g, m_g \}, c_g)$, where 
\begin{equation}
p_t = \frac{\sum_{i=1}^{H} \sum_{j=1}^{W} (i, j) \cdot m_t(i, j)}{\sum_{i=1}^{H} \sum_{j=1}^{W} m_t(i, j)}.
\end{equation}
It is worth noting that incorporating the historical observations $o_{1:t-1}$ as input is essential, especially when there is limited shared visual content between $o_t$ and $o_g$. \textit{This historical sequence acts as a smooth bridge to facilitate alignment.} 
Since the goal object represented by the mask corresponds to the target of the policy's interaction, this auxiliary task aligns the policy's actions with its visual focus, effectively improving task performance. 

\paragraph{Target Visibility Loss} 
Due to the partial observability in 3D environments, it is common for target objects in interaction trajectories to disappear from the field of view and reappear later. During such intervals, the segmentation mask for the missing object is empty. To leverage this information, we propose training the model to predict whether the target object is currently visible, formulated as: $\pi_{\text{cross}} (v_t \mid o_{1:t}, \{ o_g, m_g \})$, where $v_t$ is a binary indicator for empty segmentation masks. 
On the one hand, accurately predicting object visibility helps the policy better match the target object, avoiding a simple appearance similarity measurement between two frames. On the other hand, visibility information guides the policy to make reasonable decisions, such as confidently approaching the goal when it is visible or actively adjusting its camera to explore when the target is absent.

\paragraph{\agent Architecture} Let a training trajectory $n$ be denoted as $(c_g, \{o_{t}, m_t\}_{t=1}^{L(n)})$. A cross view index $g$ is sampled from $\{i|i\in [1, L(n)], m_i \ne \phi \}$. 
We resize all visual observations and their segmentation masks to $224 \times 224$. For encoding the visual observation $o_t$, we utilize a DINO-pretrained \citep{dino} 3-channel ViT-B/16 \citep{vit} (16 is the patch size), which outputs a token sequence of length 196, denoted as $ \{\hat{o}_t^{i}\}_{i=1}^{196} $. Inspired by \cite{dexgraspvla}, we encode the segmentation mask \( m_t \) using a 1-input-channel ViT-tiny/16, yielding $\{\hat{m}_t^{i}\}_{i=1}^{196}$. The ViT-base/16 encoder is frozen during training for efficiency, while the ViT-tiny/16 is trainable. 
To ensure spatial alignment, we fuse the cross-view condition $(o_g, m_g)$ by concatenating the feature channels: 
\begin{equation}
h_g^{i} = \text{FFN}(\text{concat}([\hat{o}_g^{i} \parallel \hat{m}_g^{i}])).
\end{equation}
Given the ability of self-attention mechanisms to capture spatial details across views, we concatenate the token sequences from two views into a sequence of length 392. A non-causal Transformer encoder module is applied \citep{transformer} for spatial fusion, obtaining a frame-level representation $x_t$:
\begin{equation}
x_t \leftarrow \text{SpatialFusion}(\{\hat{o}_t^{i}\}_{i=1}^{196}, \{h_g^{i}\}_{i=1}^{196}).
\end{equation}
Then, we leverage a causal TransformerXL \citep{transformerxl} architecture to capture temporal information among frames:
\begin{equation}
f_t \leftarrow \text{TransformerXL}(\{x_i\}_{i=1}^{t}, c_g).
\end{equation}
Finally, a light network maps $f_t$ to predict action $\hat{a}_t$, centroid $\hat{p}_t$, and visibility $\hat{v}_t$. The loss function for episode $n$ follows:
\begin{equation}
\mathcal{L}(n) = \sum_{t=1}^{L(n)} -a_t^n \log \hat{a}_t^n - p_t^n \log \hat{p}_t^n - v_t^n \log \hat{v}_t^n.
\end{equation}

\begin{figure}[t]
\begin{center}
\includegraphics[width=0.99\linewidth]{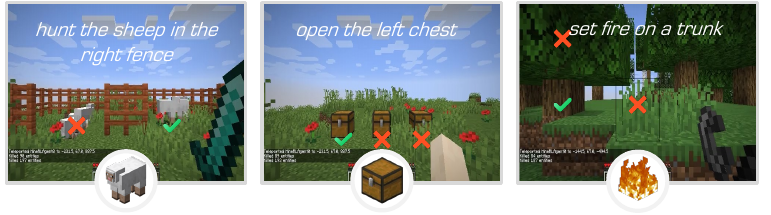}
\vspace{-3mm}
\end{center}
\caption{
\textbf{The Evaluation Metric is Spatial-Sensitive.} $\checkmark$ and $\times$ indicate the correct and incorrect objects for interaction, respectively. None of the task is trained. 
}
\label{fig:bench_demo}
\end{figure}

\begin{figure*}[htbp!]
\begin{center}
\includegraphics[width=0.99\linewidth]{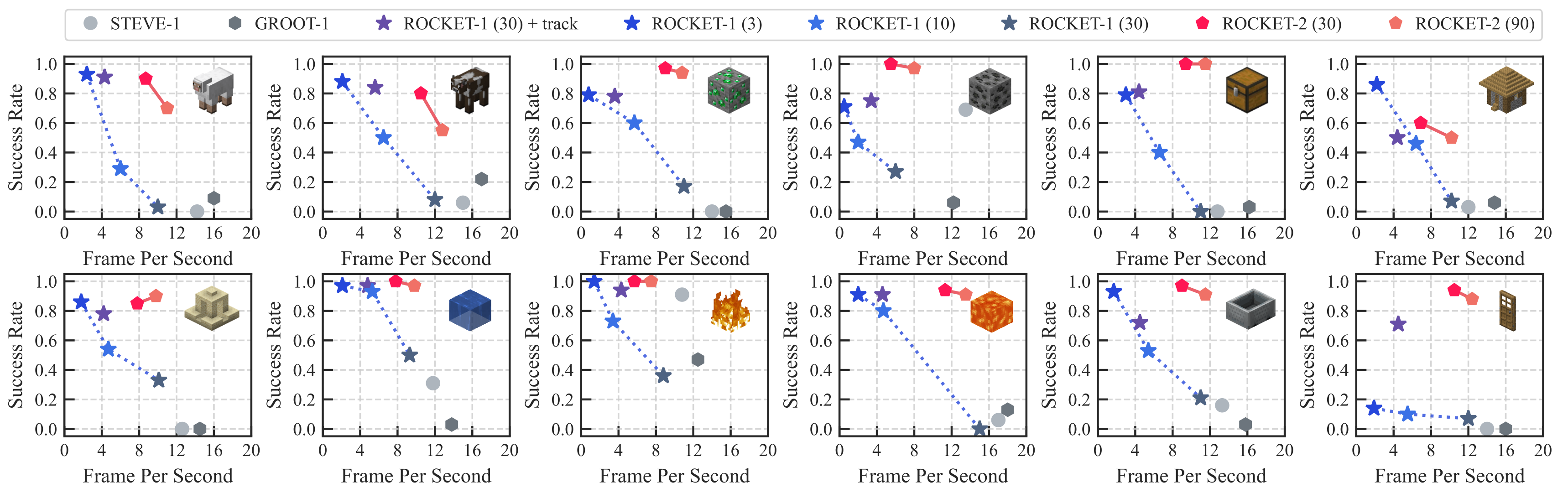}
\vspace{-3mm}
\end{center}
\caption{
\textbf{Minecraft Interaction Benchmark Results.} Our \agent achieves $3\times$ to $6\times$ faster inference, matching or surpassing ROCKET-1's peak performance in most tasks. Numbers in parentheses indicate Molmo invocation interval (steps); larger values mean higher FPS. \textit{+ track} denotes real-time SAM-2 tracking between Molmo calls. 
}
\label{fig:bench12}
\end{figure*}

\section{Experiments}
\label{sec:experiments}

We aim to address the following questions:
\textbf{(1)} How does \agent perform in terms of both accuracy and efficiency during inference?
\textbf{(2)} Can \agent follow the intent of a human from a cross-camera view in Minecraft? 
\textbf{(3)} Can \agent adapt and generalize to other 3D environments?
\textbf{(4)} How important are landmarks in cross-view goal alignment for \agent? 
\textbf{(5)} Can \agent interpret goal views from cross-episode scenarios?
\textbf{(6)} Which modules contribute effectively to training \agent? 

\subsection{Experimental Setup}
We include all the details, such as hyperparameters, benchmarks, case studies and extensive experiments, in the Appendix. 

\begin{figure}[t]
\begin{center}
\includegraphics[width=0.99\linewidth]{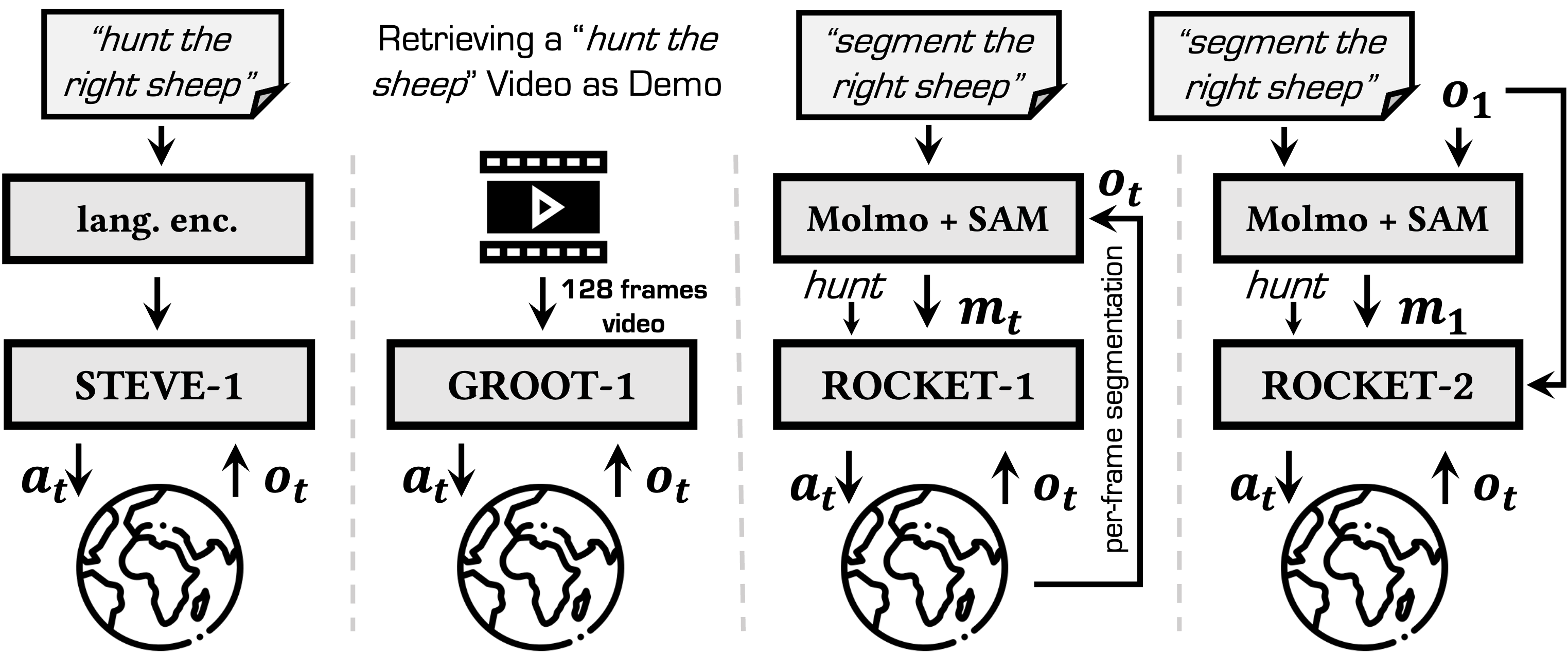}
\vspace{-2mm}
\end{center}
\caption{
\textbf{Inference Pipelines on Minecraft Interaction Benchmark.} GROOT-1 requires retreving a short video as its condition. 
ROCKET-1's segmentation is coupled with the agent's real-time observations, whereas ROCKET-2's segmentation is solely tied to the third-person view. 
}
\label{fig:infer_pipe}
\end{figure}

\paragraph{Environment and Benchmark} We use Minecraft v1.16.5 \citep{minerl, MineStudio} as the testing environment, which accepts mouse and keyboard inputs and returns a $640\times360$ RGB image per step. Following \citet{cai2024rocket}, we employ the \textit{Minecraft Interaction Benchmark} to evaluate the agent’s interaction capabilities. This benchmark includes six categories and a total of 12 tasks, covering all basic Minecraft interaction types: \textit{Hunt}, \textit{Mine}, \textit{Interact}, \textit{Navigate}, \textit{Tool}, and \textit{Place}. As this benchmark emphasizes object interaction and spatial localization, its evaluation criteria are more stringent than those in \citet{steve1} and \citet{groot-1}. We present three examples in Figure~\ref{fig:bench_demo}, more can be found in appendix. In the \textit{“hunt the sheep in the right fence”} task, success requires the agent to kill the sheep within the right fence, while killing sheep in the left fence results in failure.

\paragraph{Baselines} We compare our \agent with the following instruction-following baselines: (1) STEVE-1 \citep{steve1}: A text-conditioned agent fine-tuned from VPT \citep{vpt}, capable of solving various short-horizon tasks. 
(2) GROOT-1 \citep{groot-1}: A video-conditioned policy designed for open-ended tasks, implemented with a VAE network. 
(3) ROCKET-1 \citep{cai2024rocket}: A mask-conditioned policy capable of mastering 12 interaction tasks. Taking the \textit{hunt right sheep} task as an example, we present the inference pipelines in Figure~\ref{fig:infer_pipe}, where Molmo~\citep{molmo} and SAM~\citep{sam2} are combined to perform text-based instance segmentation. 
Given that ROCKET-2 relies on a third-person viewpoint for goal specification, in the Minecraft Interaction Benchmark, we set its third-person view to the observation acquired by the agent at the initial step. This view is then largely maintained, undergoing resets only at fixed periods (e.g., every 90 steps). This setup ensures that ROCKET-2 requires no additional privileged information. 

\begin{figure*}[ht]
\begin{center}
\includegraphics[width=0.99\linewidth]{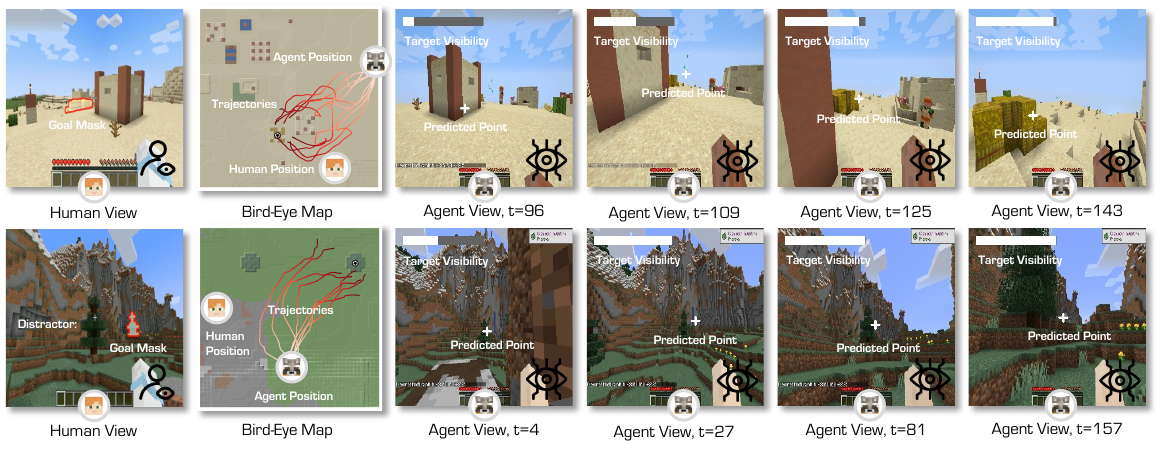}
\vspace{-4mm}
\end{center}
\caption{
\textbf{Case Study of Human-Agent Interaction.} We show how a human interacts with \agent, leveraging its spatial reasoning abilities. \textbf{(Top Row)} The human specifies a hay bale (\includegraphics[scale=0.04,valign=c]{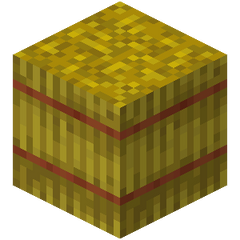}) that is not visible to \agent. By exploring the area around the visible landmark (house), \agent successfully locates the goal. \textbf{(Bottom Row)} Human specifies a target tree in the presence of a tree distractor. \agent accurately identifies the correct tree by reasoning about spatial relationships. The trajectories are visualized in BEV maps. 
}
\label{fig:human-agent}
\end{figure*}

\subsection{Performance-Efficiency Analysis}

Figure~\ref{fig:bench12} presents our Minecraft Interaction Benchmark results, showcasing various agent configurations and their performance. Specifically, ROCKET-1 (10) performs object segmentation with Molmo + SAM every 10 steps, providing no masks for the intermediate 9 steps; its "+ track" variant leverages SAM's more computationally efficient object tracking to causally generate masks for these frames. In contrast, ROCKET-2 utilizes cross-view segmentation, only requiring a single third-person view segmentation, with ROCKET-2 (90) resetting this view every 90 steps. Our initial observations highlight that STEVE-1 and GROOT-1 exhibit success rates below $20\%$ across most tasks, primarily due to their instructions' limited spatial sensitivity. While ROCKET-1 achieves over $80\%$ success with high-frequency Molmo (every 3 frames), it suffers from slow inference; lowering Molmo's frequency or enabling SAM tracking (though still expensive) significantly degrades its performance. Conversely, our ROCKET-2 agent, by decoupling goal specification from the agent's current view, autonomously tracks targets without frequent mask modifications, achieving comparable or superior performance to ROCKET-1 with a remarkable $3\times$ to $6\times$ inference speedup.

\subsection{Intuitive Human-Agent Interaction}

In Figure \ref{fig:human-agent}, we present two case studies illustrating \agent interprets human intent under the cross-view goal specification interface. 
The first case (top row) involves a task requiring the agent to approach a hay bale (\includegraphics[scale=0.04,valign=c]{figures/minecraft/hay_bale.png}) located behind a house (\includegraphics[scale=0.02,valign=c]{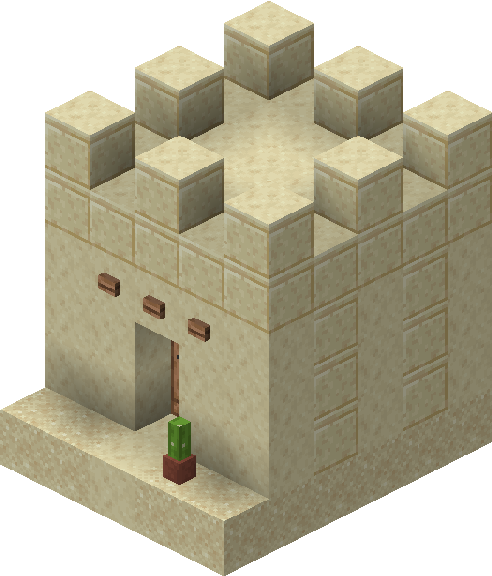}). From the human view, both the house and the hay bale are visible, whereas \agent initially observes only the house. A key challenge arises from the differing camera views: the human and \agent perceive the scene from opposite sides of the house. To analyze the agent’s behavior, we visualize both its camera views and its trajectories on a bird’s-eye map. We observe that \agent effectively infers the hay bale’s potential location and successfully navigates toward it. This is reflected in the increasing target visibility score and the movement of the predicted point. Interestingly, the bird’s-eye view reveals that \agent approaches the target from both sides of the house, demonstrating diversity in route selection. 
The second case (bottom row) showcases \agent’s ability to distinguish between a distractor and the human-specified goal object, despite their visual similarity. It highlights that agent's spatial reasoning extends beyond object appearance.

\begin{figure*}[ht]
\begin{center}
\includegraphics[width=0.99\linewidth]{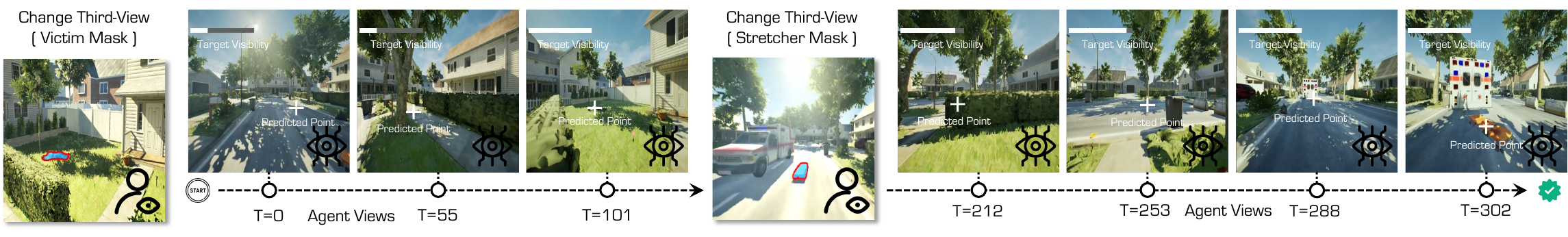}
\vspace{-3mm}
\end{center}
\caption{
\textbf{Zero-Shot 3D Environment Generalization.} \agent can find and transport the victim to the stretcher in the unseen \textbf{Unreal} Engine. We map the \textit{left click}/\textit{right click} actions of Minecraft to the \textit{carry up}/\textit{put down} actions of Unreal to interact with objects, respectively. 
}
\label{fig:zero-shot}
\end{figure*}

\subsection{Zero-Shot 3D Worlds Generalization}
We demonstrate that by mapping the action space of Minecraft to unseen 3D games (refer to Table~\ref{tab:action_mapping}), \agent achieves zero-shot generalization, despite \textbf{being trained solely on the Minecraft dataset}. We attribute this to two key factors: (1) \textit{the DINO-pretrained ViT backbone is frozen during training, preserving its general 3D view perception capability}; and (2) \textit{the learned cross-view goal alignment skill is domain-agnostic}. We evaluated \agent in the ATEC 2025 AI and Robotics Challenge~\footnote{https://github.com/atecup/atec2025\_software\_algorithm}, which is built in the Unreal Engine and assigns agents to \textit{locate injured people and transport them to stretchers}. The benchmark provides both textual (a long text description) and visual cues (a third-person view) about the locations of the victims. 
Note that, \agent takes in \textbf{solely visual cues} (Figure~\ref{fig:zero-shot}). As shown in Table~\ref{tab:atec2025}, \agent outperforms the \textbf{strong baseline} provided by the organizers, a Gemma-3 and YOLO-based two-tiered decision system using both textual and visual cues, by $12\%$, despite not being fine-tuned on the Unreal. 
Furthermore, the Unreal environment provides observations at a $640\times480$ resolution, a notable deviation from the $640\times360$ resolution of the training dataset. This robustly substantiates that cross-view goal alignment enables the acquisition of transferable 3D decision representations within one environment and their successful migration to another. Such capabilities underscore the significant feasibility of developing general-purpose multi-task agents for diverse 3D environments. 

\begin{table}[h]
\setlength{\tabcolsep}{1.0 mm}
\centering
\caption{ \textbf{Bridging the Minecraft Action Space and Other 3D Games.} ``/'' denotes the masked action. } \label{tab:action_mapping} 
\renewcommand{\arraystretch}{1.0}
\footnotesize
\begin{adjustbox}{width=\linewidth}
\begin{tabular}{@{}llll@{}}
\toprule
\textbf{Minecraft} & \textbf{DeepMind Lab} & \textbf{Doom} & \textbf{Unreal} \\ \midrule
$\text{forward}=1$ & $a[3]=1$     & $\text{discrete}[2]=1$   & $\text{velocity}=+100$  \\
$\text{back}=1$    & $a[3]=-1$    & $\text{discrete}[3]=1$   & $\text{velocity}=-100$ \\
$\text{attack}=1$  & $a[4]=1$     & $\text{discrete}[1]=1$   & $\text{pick}=1$  \\
$\cdots$   & $\cdots$     & $\cdots$          & $\cdots$             \\
$\text{yaw}=x$     & $a[0]=4.75x$ & $\text{continuous}=x$ & $\text{angular}=x$     \\
$\text{pitch}=x$   & $a[1]=2.78x$ & /          & $\text{viewport}=x$             \\ 
\bottomrule
\end{tabular}
\end{adjustbox}
\end{table}

\begin{table}[h]
\setlength{\tabcolsep}{1.5 mm}
\centering
\caption{\textbf{Results on ATEC 2025 Robotics Challenge}.} \label{tab:atec2025} 
\renewcommand{\arraystretch}{1.0}
\footnotesize
\begin{adjustbox}{width=\linewidth}
\begin{tabular}{@{}lcc@{}}
\toprule
\textbf{Agent}         & \agent (Zero-Shot)   & YOLO-Gemma 3    \\ \midrule
\textbf{Rank}          & $\mathbf{1/57}$      & $20/57$  \\ 
\textbf{Success Rate}  & $\boldsymbol{38}\%$  & $26\%$  \\ \bottomrule 
\end{tabular}
\end{adjustbox}
\vspace{-3mm}
\end{table}

\begin{figure*}[ht]
\begin{center}
\includegraphics[width=0.99\linewidth]{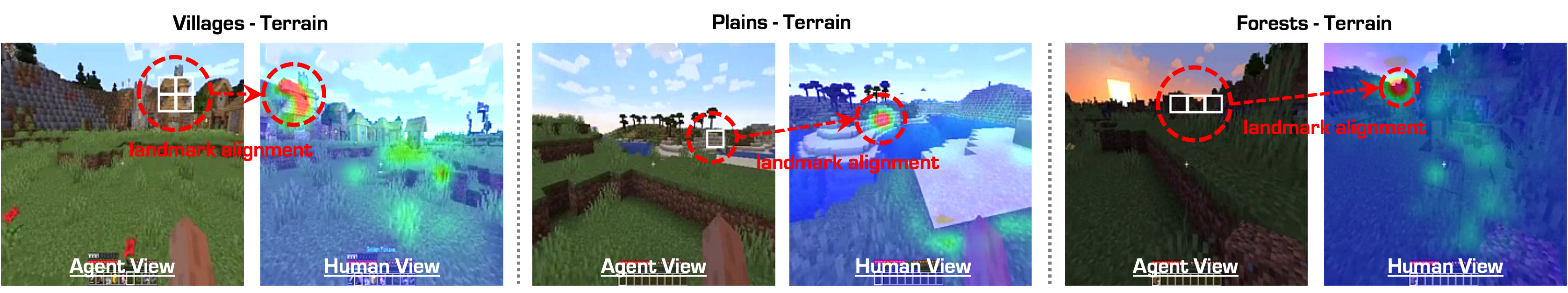}
\end{center}
\vspace{-2mm}
\caption{
\textbf{Visualization Analysis of Landmarks' Alignment.} The vision patches (identified by white grid) represent a chosen background landmark in the agent’s current view (instead of the goal object). We generate an attention map with the \textbf{spatial fusion transformer} using these patches as queries and the human view patches as keys and values. We find that \agent well aligns selected landmarks across views. 
}
\label{fig:attention}
\end{figure*}

\subsection{Ablation Studies on Auxiliary Objectives}

To evaluate the impact of auxiliary losses on model performance, we define three variants: (1) only \textit{behavior cloning loss}, (2) + \textit{target visibility loss}, and (3) the full version with + \textit{cross-view consistency loss}. We conduct experiments on three tasks: \textit{Navigate to House in a Village}(\includegraphics[scale=0.40,valign=c]{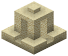}), \textit{Mine Emerald}(\includegraphics[scale=0.40,valign=c]{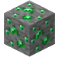}), and \textit{Interact with the Left Chest}(\includegraphics[scale=0.40,valign=c]{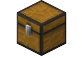}). We find that the BC-only variant achieves an average success rate of only $65\%$, demonstrating that the action signal is insufficient for learning spatial alignment. Adding \textit{target visibility loss} improves performance by $6\%$, while further incorporating \textit{cross-view consistency loss} boosts the success rate to $94\%$. This proves that leveraging temporal consistency and introducing vision-based auxiliary losses can greatly enhance cross-view goal alignment and inference-time decision-making. 

\begin{table}[ht]
\footnotesize
\setlength{\tabcolsep}{2.0 mm}
\renewcommand{\arraystretch}{1.1}
\caption{
\textbf{Ablation Study on Auxiliary Objectives.} The final loss function for each row is the cumulative sum of all loss functions from the preceding ones. 
} \label{tab:ablation-objectives}
\begin{adjustbox}{width=\linewidth}
\begin{tabular}{@{}lcccc@{}}
\toprule
\textbf{Model Variants}     & \includegraphics[scale=0.40,valign=c]{figures/icons/navigate_house.png} & \includegraphics[scale=0.40,valign=c]{figures/icons/mine_emerald.png} & \includegraphics[scale=0.40,valign=c]{figures/icons/interact_chest.png} & \textbf{Avg.} \\ \midrule
behavior cloning         & $0.52$ & $0.78$  & $0.65$ & $0.65$  \\
+ target visibility      & $0.63$ & $0.83$  & $0.68$ & $0.71$  \\
+ cross-view consistency & $\mathbf{0.85}$ & $\mathbf{0.97}$  & $\mathbf{1.00}$ & $\mathbf{0.94}$  \\ \bottomrule
\end{tabular}
\end{adjustbox}
\vspace{-3mm}
\end{table}

\subsection{Landmarks Attention Visualization}

Prominent non-goal objects, referred to as ``landmarks'', play a crucial role in assisting humans or agents in localizing goal objects within a scene. For instance, when multiple objects with similar appearances are present, spatial relationships between the goal and landmarks can aid in distinction. In this subsection, we aim to explore whether \agent implicitly learns landmark alignment by visualizing the attention weights of its spatial transformer. 

Specifically, we prepare a current view observation and a third view with goal segmentation. Before being fed into the spatial transformer, both views are encoded into $14 \times 14 = 196$ tokens: $\{\hat{o}_t^{i}\}_{i=1}^{196}$ and $\{h_g^{i}\}_{i=1}^{196}$ (notations are consistent with Sec. \ref{sec:method}). We inspect the softmax-normalized attention map of the first self-attention layer in the spatial transformer, denoted as $\{a_{i,j}\}_{i,j=1}^{392}$, where $a_{i, 197:392}$ represents the attention map generated by using patch $i$ from the current view as the query and all patches from the third view as keys and values. This map is overlaid on the third view (goal view) to reflect its responsiveness to patch $i$ in the current view. Since landmarks may span multiple patches, we aggregate the response maps of different patches to form the final attention map $\{m_i\}_{i=1}^{196}$:
$m_{i} = \frac{1}{|L|} \sum_{x \in L} a_{x, i+196}$,
where $L$ denotes the set of patches in the current view representing a specific landmark. Notably, the selected landmarks do not overlap with the goal segmentation. 
As shown in Figure \ref{fig:attention}, we present three sets of data covering \textit{villages}, \textit{plains}, and \textit{forest} terrains. In the left plot, the white grid indicates the selected landmark patches, while the right plot shows the human-view attention response to the chosen landmarks. Our findings reveal that \agent effectively matches cross-view consistency even under significant geometric deformations and distance variations. Surprisingly, in the last data point, even subtle forest depressions are accurately matched.

\begin{figure}[t]
\begin{center}
\includegraphics[width=0.99\linewidth]{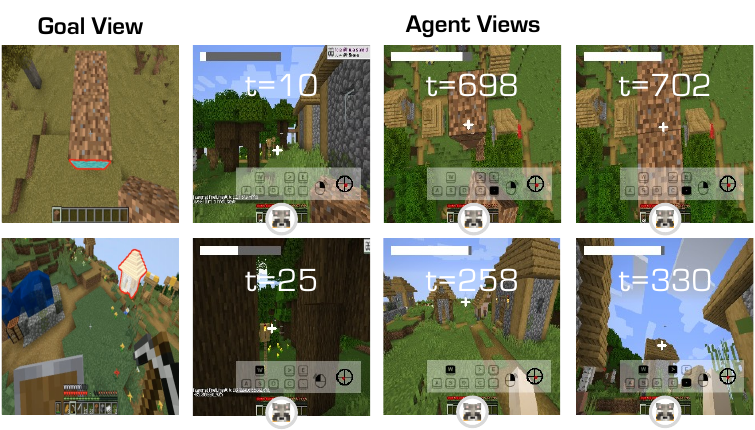}
\end{center}
\vspace{-3mm}
\caption{
\textbf{Cross-Episode Generalization.} 
The goal view does not exist within the agent’s world but originates from a different episode. \agent attempts to infer the semantic intent underlying the goal specification (\textit{building a bridge} and \textit{approaching a house}). 
}
\label{fig:generalization}
\end{figure}

\subsection{Cross-Episode Cross-View Goal Alignment}
We observe that \agent exhibits cross-episode generalization capabilities. As shown in Figure \ref{fig:generalization}, the selected goal views come from different episodes, each generated with a unique world seed. In the top-row example, the goal view is from a “bridge-building” episode set in the \textit{savanna} biome, where the player is placing a dirt block to build the bridge. After feeding forward the goal view, we place \agent in a \textit{forest} biome and observe its behavior. We find that it first exhibits pillar-jumping behavior, and after placing many blocks, it begins to build the bridge horizontally. Although it ultimately failed to build the perfect bridge, the emergent behavior still indicates that \agent attempts to understand the underlying semantic information when there is no landmark match across views. In the bottom row, the goal view is taken from a Minecraft creative mode, observing a house from the sky— a view never seen during training. We find that \agent explores its environment and successfully identifies a visually similar house. This demonstrates \agent’s robustness to a variety of goal views.

\section{Conclusion}
\label{sec:conclusion}

We propose a cross-view goal specification method to improve human-agent interaction in embodied worlds. To address misalignment between agent and human views, we introduce cross-view consistency and visibility losses. Our agent sets a new benchmark on the \textit{Minecraft Interaction Benchmark}, achieving $3{\times}$–$6{\times}$ higher efficiency. Visualizations support the effectiveness of our method. Our zero-shot generalization results underscore the promise of cross-view goal alignment as a core for building general-purpose decision-making agents in 3D worlds. 

\section*{Acknoledgements}
This work was supported by the National Science and Technology Major Project \#2022ZD0114902. We sincerely appreciate their generous support, which enabled us to conduct this research.

\bibliography{refs}
\citestyle

\newpage
\appendix

\twocolumn[{
\begin{center}
    \LARGE \textbf{Appendix} \\[0.5ex]
    \large Implementation Details and Extended Results
\end{center}
\vspace{1.5em}
}]

\begin{figure*}[ht]
\begin{center}
\includegraphics[width=0.99\linewidth]{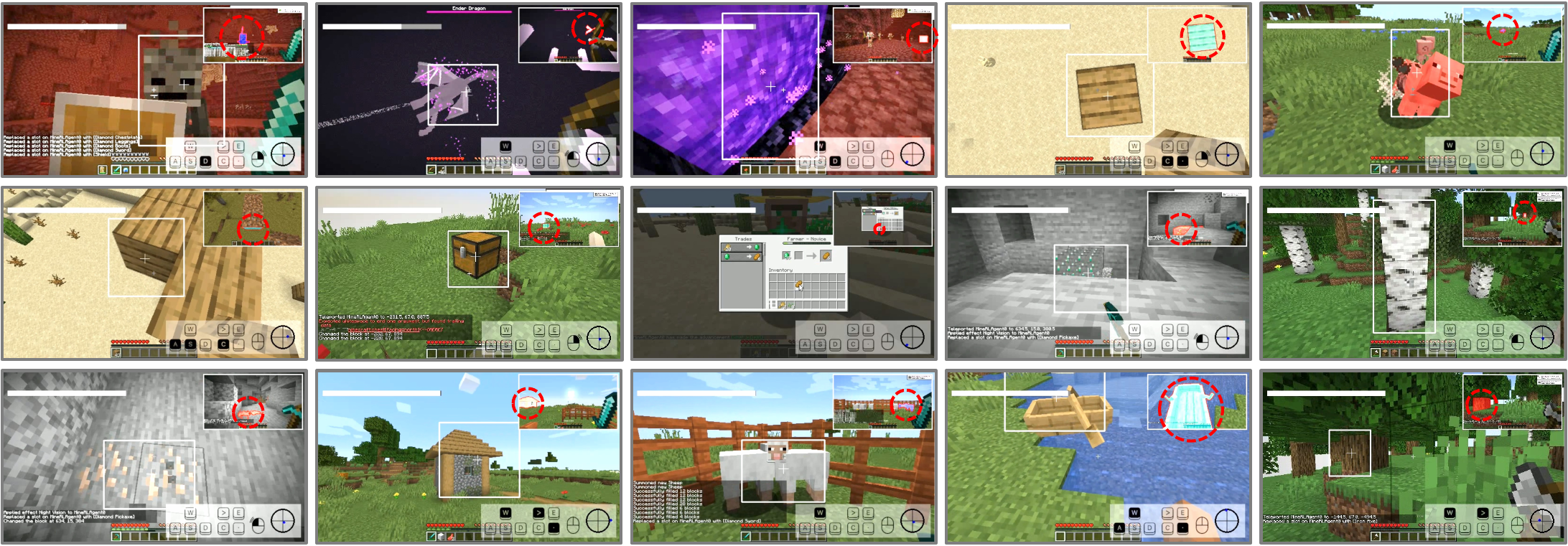}
\end{center}
\caption{\textbf{Minecraft Demonstrations.} ROCKET-2 showcases a diverse range of skills, including combat with monsters, locating the Ender portal, bridge construction, ore mining, and crafting. Cross-view perspectives with segmented and circled masks are displayed in the top right window. Predicted bounding boxes, visibility, and actions are overlaid on the images, highlighting the agent’s robust 3D visual goal understanding and alignment with human-specified targets.}
\label{fig:minecraft-demo}
\end{figure*}

\section{Environments}

ROCKET-2 is trained in Minecraft~\citep{MineStudio} and evaluated not only within Minecraft but also across a diverse set of \textbf{out-of-domain} environments, including Unreal Engine~\citep{unrealzoo}, Doom~\citep{vizdoom}, DeepMind Lab~\citep{dmlab}, and several commercial Steam games. \Cref{tab:environments_overview} provides an overview of their observation and action spaces. To enable cross-environment testing, we construct a rule-based mapping from Minecraft’s action space to those of the target environments (\cref{tab:action_mapping}), without relying on environment-specific fine-tuning or carefully crafted adaptations.

\begin{figure*}[ht]
\begin{center}
\includegraphics[width=0.99\linewidth]{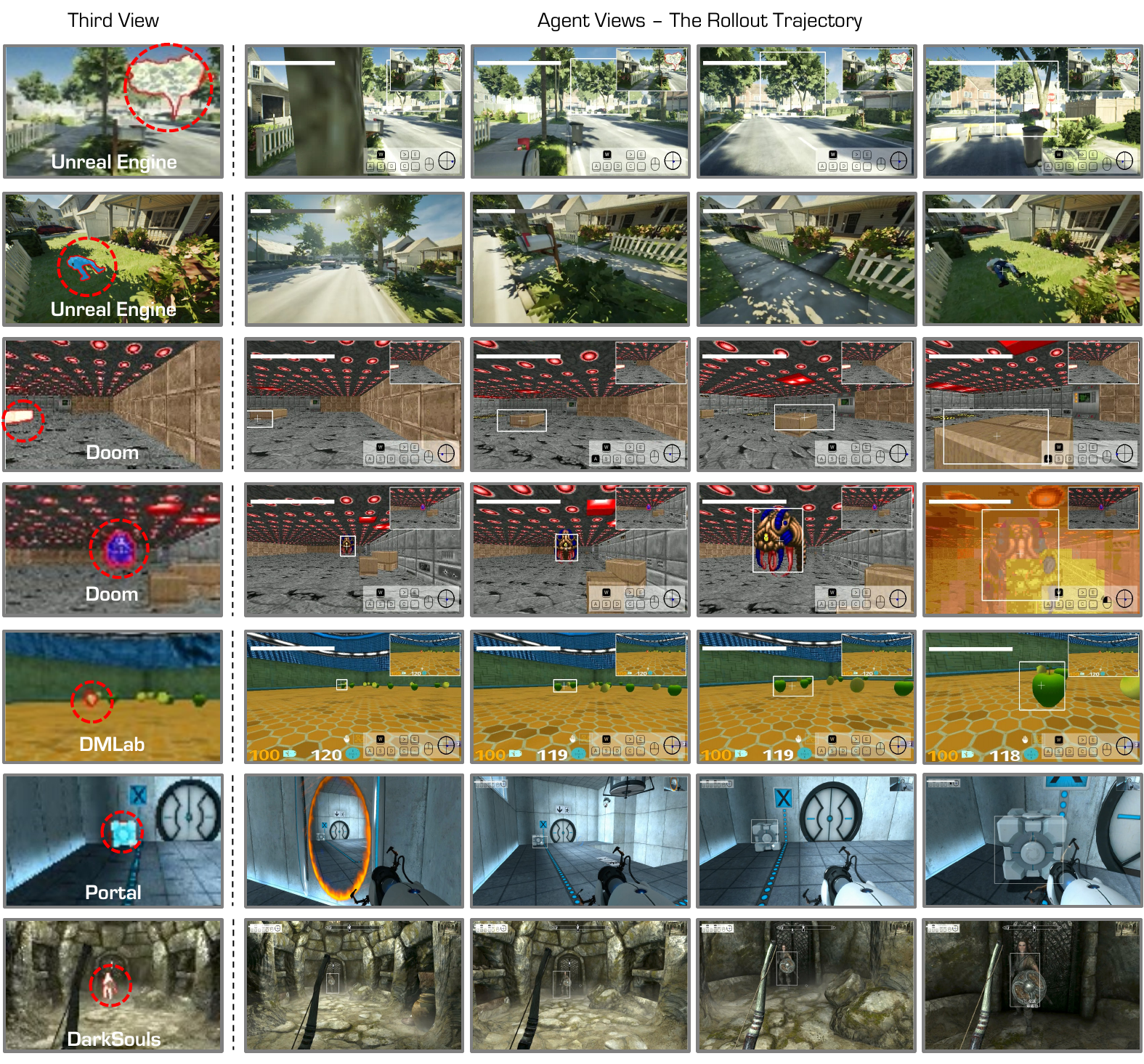}
\end{center}
\caption{
\textbf{Zero-Shot Demonstrations.}
We evaluate ROCKET-2 in a variety of previously unseen environments, including Unreal Engine~\citep{unrealzoo}, Doom~\citep{vizdoom}, DeepMind Lab~\citep{dmlab}, Portal, and Dark Souls. Remarkably, ROCKET-2 exhibits strong exploration abilities guided by visual cues in Unreal Engine, successfully transfers certain skills such as attacking in Doom, and maintains robustness to observations that differ significantly from those in Minecraft (Unreal Engine, Portal, and Dark Souls).
}
\label{fig:zs-demo}
\end{figure*}

We present qualitative demonstrations in both Minecraft and unseen environments in Figure~\ref{fig:minecraft-demo} and Figure~\ref{fig:zs-demo}, where cross-view images with human-provided masks serve as goal specifications. In Minecraft, ROCKET-2 showcases advanced interactive abilities, including defeating the Ender Dragon and constructing long-span bridges—tasks previously unattainable by existing agents. Remarkably, ROCKET-2 also exhibits strong generalization to unfamiliar environments: it performs visually-guided exploration in Unreal Engine, effectively transfers combat behaviors to Doom, and remains robust despite drastic differences in observation styles across games such as Portal and Dark Souls. These results highlight the model’s ability to leverage visual cues and make meaningful decisions in a zero-shot setting, underscoring the versatility and generality of cross-view goal alignment.

\definecolor{tableheadergray}{HTML}{EFEFEF}
\definecolor{tablelineblue}{HTML}{0057B8}
\definecolor{grey}{rgb}{0.5,0.5,0.5}
\definecolor{trainingenvcolor}{HTML}{F0F8FF}

\begin{table*}[ht!]
    \centering
    \caption{\textbf{Overview of Environments.} The \textbf{\textcolor{tablelineblue}{blue line}} separates the training environment from the zero-shot evaluation environments. Minecraft is our primary training environment, while others serve for zero-shot generalization assessment.}
    \label{tab:environments_overview}
    \renewcommand{\arraystretch}{1.5} 

    \small
    \begin{tblr}{
        colspec = { Q[m, l, wd=3.2cm, font=\bfseries] X[l, m] X[l, m] },
        row{1} = {bg=tableheadergray, fg=black, font=\bfseries}, 
        row{2} = {bg=trainingenvcolor}, 
        row{even[4-Z]} = {bg=tableheadergray},
        cell{2-Z}{1} = {cmd=\small}, 
        measure = vbox,
    }
        \toprule
        Environment & Observation Space Details & Action Space Details \\
        \midrule
        \textbf{Minecraft} \textcolor{grey}{(Training)} \par \citep{MineStudio} &
        \begin{itemize}[nosep, leftmargin=*, topsep=0pt, partopsep=0pt, itemsep=2pt]
            \item \textbf{Visual}: First-person RGB images ($640 \times 360$).
            \item \textbf{Numerical}: Comprehensive player information (e.g., health, food levels); not utilized during training and inference.
        \end{itemize} &
        \begin{itemize}[nosep, leftmargin=*, topsep=0pt, partopsep=0pt, itemsep=2pt]
            \item \textbf{Discrete} actions: \texttt{attack}, \texttt{back}, \texttt{forward}, \texttt{hotbar}, \texttt{inventory}, \texttt{jump}, \texttt{left}, \texttt{right}, \texttt{sneak}, \texttt{sprint} and \texttt{use}.
            \item \textbf{Continuous} actions: adjusting \textit{pitch} and \textit{yaw} in $[-180, 180]$.
            \item For the output of agent, we take the hierarchical action space (\citet{vpt}) of 8,641 discrete button combinations and 121 camera movement bins.
        \end{itemize} \\
        \hline[\heavyrulewidth, tablelineblue]
        \textbf{Unreal Engine} \par \citep{unrealzoo} &
        \begin{itemize}[nosep, leftmargin=*, topsep=0pt, partopsep=0pt, itemsep=2pt]
            \item \textbf{Visual}: First-person RGB images as well as goal images ($640 \times 480$). The observation resolution notably differs from that of Minecraft.
            \item \textbf{Goal}: The agent must rescue objects and transport them to designated stretchers, requiring exploration of complex 3D environments guided by visual goal cues.
        \end{itemize} &
        \begin{itemize}[nosep, leftmargin=*, topsep=0pt, partopsep=0pt, itemsep=2pt]
            \item \textbf{Discrete} actions: \texttt{carry}, \texttt{drop}, \texttt{jump}, \texttt{crouch}, and \texttt{open the door}.
            \item \textbf{Continuous} actions: adjusting angular velocity (turning), linear velocity (moving forward/backward), and viewpoint (masked).
        \end{itemize} \\
        \textbf{ViZDoom} \par \citep{vizdoom} &
        \begin{itemize}[nosep, leftmargin=*, topsep=0pt, partopsep=0pt, itemsep=2pt]
            \item \textbf{Visual}: First-person RGB view (e.g., $320 \times 240$).
            \item We use the basic config for demonstration.
        \end{itemize} &
        \begin{itemize}[nosep, leftmargin=*, topsep=0pt, partopsep=0pt, itemsep=2pt]
            \item 3 primary actions are enabled in the basic setting: \texttt{move left/right}, \texttt{shoot (attack)}.
        \end{itemize} \\
        \textbf{DeepMind Lab} \par \citep{dmlab} &
        \begin{itemize}[nosep, leftmargin=*, topsep=0pt, partopsep=0pt, itemsep=2pt]
            \item \textbf{Visual}: First-person RGB images ($640 \times 480$).
            \item We choose map \texttt{seekavoid\_arena\_01}, a fruit-gathering environment with navigation scenarios.
        \end{itemize} &
        \begin{itemize}[nosep, leftmargin=*, topsep=0pt, partopsep=0pt, itemsep=2pt]
            \item \textbf{Discrete} actions: \texttt{move back/forward}, \texttt{fire}, \texttt{jump}, \texttt{crouch}, and \texttt{strafe left/right}.
            \item \textbf{Continuous} actions: adjusting angular velocity (turning), and viewpoint (masked) in angular per second of game time.
        \end{itemize} \\
        \textbf{Steam Games} \par (Portal/Dark Souls) &
        \begin{itemize}[nosep, leftmargin=*, topsep=0pt, partopsep=0pt, itemsep=2pt]
            \item \textbf{Visual}: First-person RGB view, resized to $640 \times 360$.
        \end{itemize} &
        \begin{itemize}[nosep, leftmargin=*, topsep=0pt, partopsep=0pt, itemsep=2pt]
            \item Same as Minecraft, where output actions are mapped to human keyboard and mouse movements.
        \end{itemize} \\ 
        \bottomrule
    \end{tblr}
\end{table*}

\section{Implementation Details} \label{sec:appendix-impl}

We present the model architecture, hyperparameters, and optimizer configurations of \agent in \cref{table:implementation}. During training, each trajectory is divided into segments of length 128 to reduce memory requirements. We initialize the view backbone that is used to encode RGB images with DINO weights and freeze it for training efficiency. 
During inference, \agent can access up to 128 key-value attention caches of past observations. Most training parameters follow those from prior works such as \cite{ cai2024rocket, vpt}.

\begin{table}[H]
\setlength{\tabcolsep}{5.0 mm}
\centering
\renewcommand{\arraystretch}{1.1}
\caption{
\textbf{Detailed Training Hyperparameters.}
} \label{table:implementation}
\begin{adjustbox}{width=\linewidth}
\begin{tabular}{@{}lc@{}}
\toprule
\textbf{Hyperparameter} & \textbf{Value} \\ \midrule
Input Image Size & $224 \times 224$ \\
Hidden Dimension & $1024$\\
View Backbone & ViT-base/16\ \ (DINO-v1)\\
Mask Backbone& ViT-tiny/16 (1-channel) \\
Spatial Transformer& PyTorch Transformer \\
Number of Spatial Blocks & $4$\\
Temporal Transformer & TransformerXL \\
Number of Temporal Blocks & $4$\\
Trajectory Chunk size& $128$\\
Optimizer& AdamW\\
Learning Rate  & $0.00004$\\
\bottomrule
\end{tabular}
\end{adjustbox}
\end{table}

\section{Training Datasets}

We train ROCKET-2 using 1.6 billion frames of human gameplay data collected by OpenAI~\citep{vpt}. Each frame is annotated with fine-grained interaction events, including actions such as mining, crafting, item usage, and combat. This large-scale dataset provides rich behavioral diversity across a wide range of tasks and environmental contexts.  
Building on the protocol established by ROCKET-1~\citep{cai2024rocket}, we further enrich each interaction event with frame-level object masks that identify the involved entities or objects. For each interaction event, we ensure that all associated masks correspond to the same target entity. From these, one masked frame is randomly sampled to serve as the cross-view goal specification for training. This process enables the model to learn to align its current egocentric observation with high-level goals presented from different viewpoints.  
The interaction types used in training include \textit{use}, \textit{break}, \textit{approach}, \textit{craft}, and \textit{kill entity}, following the taxonomy introduced in ROCKET-1. These categories cover a broad spectrum of interaction patterns, equipping ROCKET-2 with the ability to generalize across diverse task structures and object affordances. 

\begin{figure*}[ht]
\begin{center}
\includegraphics[width=0.99\linewidth]{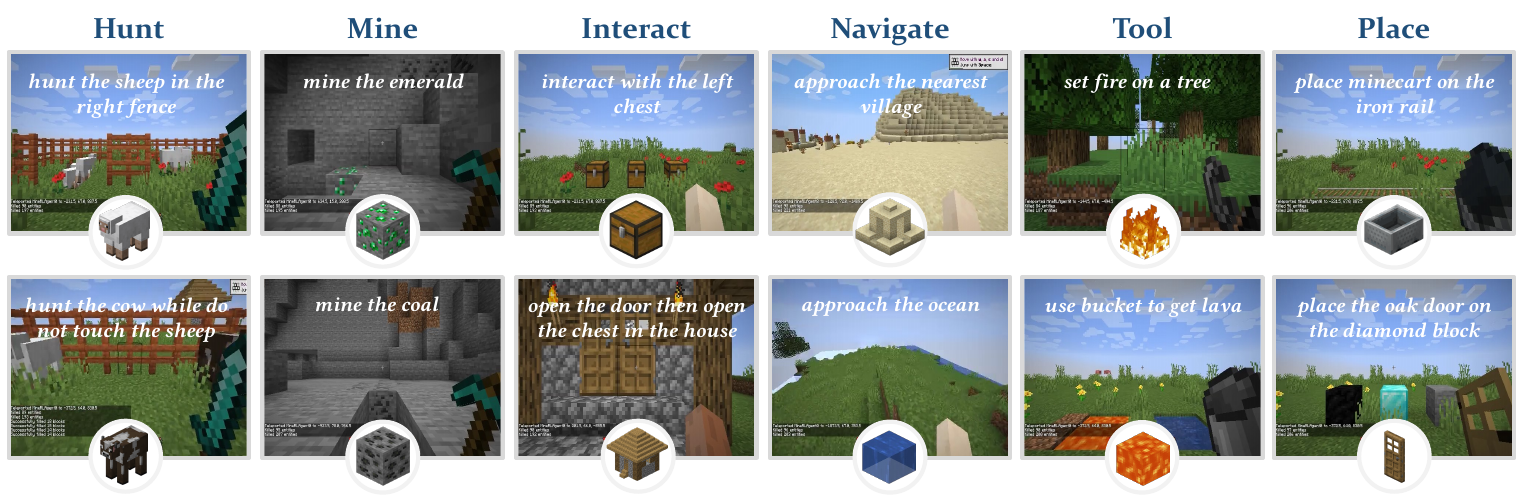}
\end{center}
\caption{
        \textbf{Minecraft Interaction Benchmark.} The benchmark in ROCKET-1 is designed to test an agent's interaction capabilities with visual goals. The 12 tasks challenge the agent's ability to follow visual cues to interact with objects at precise spatial locations, a key advancement over previous benchmarks. Image credit: \citet{cai2024rocket}.
}
\label{fig:bench_complete}
\end{figure*}

\section{Minecraft Interaction Benchmark} \label{sec:bench}

To assess ROCKET-2’s ability to accurately interact with target objects under cross-view settings, we adopt the \textbf{Minecraft Interaction Benchmark} introduced in ROCKET-1. This benchmark consists of 12 tasks grouped into six categories: \textit{Hunt}, \textit{Mine}, \textit{Interact}, \textit{Navigate}, \textit{Tool}, and \textit{Place}. Each task requires the agent to identify and interact with specific objects or locations based on visual-temporal prompts. An overview of the benchmark is shown in \cref{fig:bench_complete}. 
Compared to prior benchmarks such as \citet{mcu}, this benchmark places a stronger emphasis on precise spatial interaction—requiring the agent not just to recognize object categories, but to distinguish between multiple instances of the same object type and interact with the correct one at a particular location. This setup significantly raises the bar for visual grounding and spatial understanding, making it a more rigorous testbed for evaluating cross-view interaction capabilities in partially observable 3D environments. 

\section{Analyzing Failure Cases}
We conduct a detailed analysis of failure cases during \agent's task execution and identify three primary sources of error:

\paragraph{Prediction Drift} 
When pursuing distant targets over extended sequences, the predicted interaction point gradually drifts away from the actual object. This issue arises because the agent relies heavily on temporal consistency from memory to maintain object identity. However, during training, the model was only exposed to sequences of up to 128 steps. As a result, it struggles to generalize to longer horizons at test time, revealing limitations in its long-range memory modeling.

\paragraph{Distance Perception Error} 
In cases where the agent's current egocentric view differs significantly from the goal-provided view, we observe that the agent sometimes stops one step short of the target, leading to failed interactions. This suggests a misjudgment of spatial proximity under high viewpoint discrepancy. In particular, updating the goal to reflect the current view of the agent mitigates this problem, indicating that the model encounters larger cross-view gaps during inference than during training, affecting its precision of spatial reasoning. 

\paragraph{Action Jitter} 
When evaluating the original version of \agent, we notice substantial action jitter—rapid, unstable changes in predicted actions—which often results in failures for tasks requiring fine control, such as block placement. We find that feeding previous actions as input during both training and inference significantly improves action stability, leading to smoother and more reliable behavior. 

\section{Comparison with ROCKET-1}
ROCKET-1 tackles interaction problems in 3D worlds by training a visuomotor policy to identify interaction targets based on semantic segmentations in the visual context. While it resolves traditional goal images' ambiguity and generation challenges, it relies on SAM-2 \citep{sam2} to track goals and segment them during inference, severely limiting real-time performance. 
Additionally, its frame-by-frame segmentation training fails to build the policy's 3D spatial perception, making it heavily dependent on SAM-2's tracking capabilities. 
\agent is a cross-view segmentation-conditioned policy that. Unlike ROCKET-1, it enables the policy to align the goal across camera views by itself and removes the need for real-time segmentation. 

\section{Hindsight Trajectory Relabeling} 
There are two main approaches to collecting labeled trajectory data: (1) providing instructions for contractors to collect trajectories in real-time, ensuring a causal link between actions and labels but incurring high costs and limited scalability; and (2) gathering large amounts of trajectories and generating labels through post-processing, known as \textit{hindsight trajectory relabeling}. While the first approach \citep{language_table, rt-x} produces higher-quality data, its cost constraints have led most research to adopt the second, more scalable one. 
\cite{her} was the first to reinterpret a trajectory's behavior using its final frame, greatly improving data utilization and inspiring subsequent research on goal-conditioned policies. \cite{steve1} extended the approach by using the last 16 frames of a trajectory as the more expressive goal. \cite{rt-sketch} introduced hand-drawn sketches as a goal modality, reducing semantic ambiguity. \cite{rt-trajectory} converted the robotic end-effector moving sketch to a 2D image as the goal, providing richer procedural details. ROCKET-1 introduced backward trajectory relabeling, which first identifies interaction objects and events, then utilizes object tracking models \citep{sam2} to generate frame-level segmentations. This data supports training segmentation-conditioned policies. Our paper explores using this dataset to train policies for cross-view goal alignment. 

\begin{table}[H]
\renewcommand{\arraystretch}{1.1}
\caption{
\textbf{ROCKETs Inference Pipeline Formulation.} Molmo can locate the target object based on the task prompt. SAM uses the point to generate object mask $m_t$ \wrt $o_t$ and supports per-frame tracking.
} \label{tab:inference-pipeline}
\begin{adjustbox}{width=\linewidth}
\begin{tabular}{@{}lc@{}}
\toprule
\textbf{Model}  & \textbf{Inference Pipeline} \\ \midrule
R1(3)           & \makecell{
            $\mathbf{m_1} \leftarrow \text{SAM}(o_1, \text{Molmo}(o_1, \text{prompt}))$ \\ 
            $\pi_{R1}(a_t | o_1, \mathbf{m_1}, o_2, o_3, o_4, \mathbf{m_4}, o_5, o_6, o_7, \mathbf{m_7}, \cdots)$ }         \\ \midrule
R1(30)+track   & \makecell{
            $\mathbf{m_{1:30}} \leftarrow \text{SAM}(o_{1:30}, \text{Molmo}(o_1, \text{prompt}))$ \\ 
            $\pi_{R1}(a_t | o_1, \mathbf{m_1}, o_2, \mathbf{m_2}, o_3, \mathbf{m_3}, o_4, \mathbf{m_4}, \cdots)$ }         \\ \midrule
\textbf{R2(60)}& \makecell{
            $\mathbf{m_{1}} \leftarrow \text{SAM}(o_{1}, \text{Molmo}(o_1, \text{prompt}))$ \\ 
            $\pi_{R2}(a_t | o_1, \mathbf{m_1}, o_2, o_3, o_4, \cdots, o_{60}, o_{61}, \mathbf{m_{61}}, \cdots)$ }         \\ \bottomrule
\end{tabular}
\end{adjustbox}
\end{table}

\begin{figure*}[ht]
\begin{center}
\includegraphics[width=0.99\linewidth]{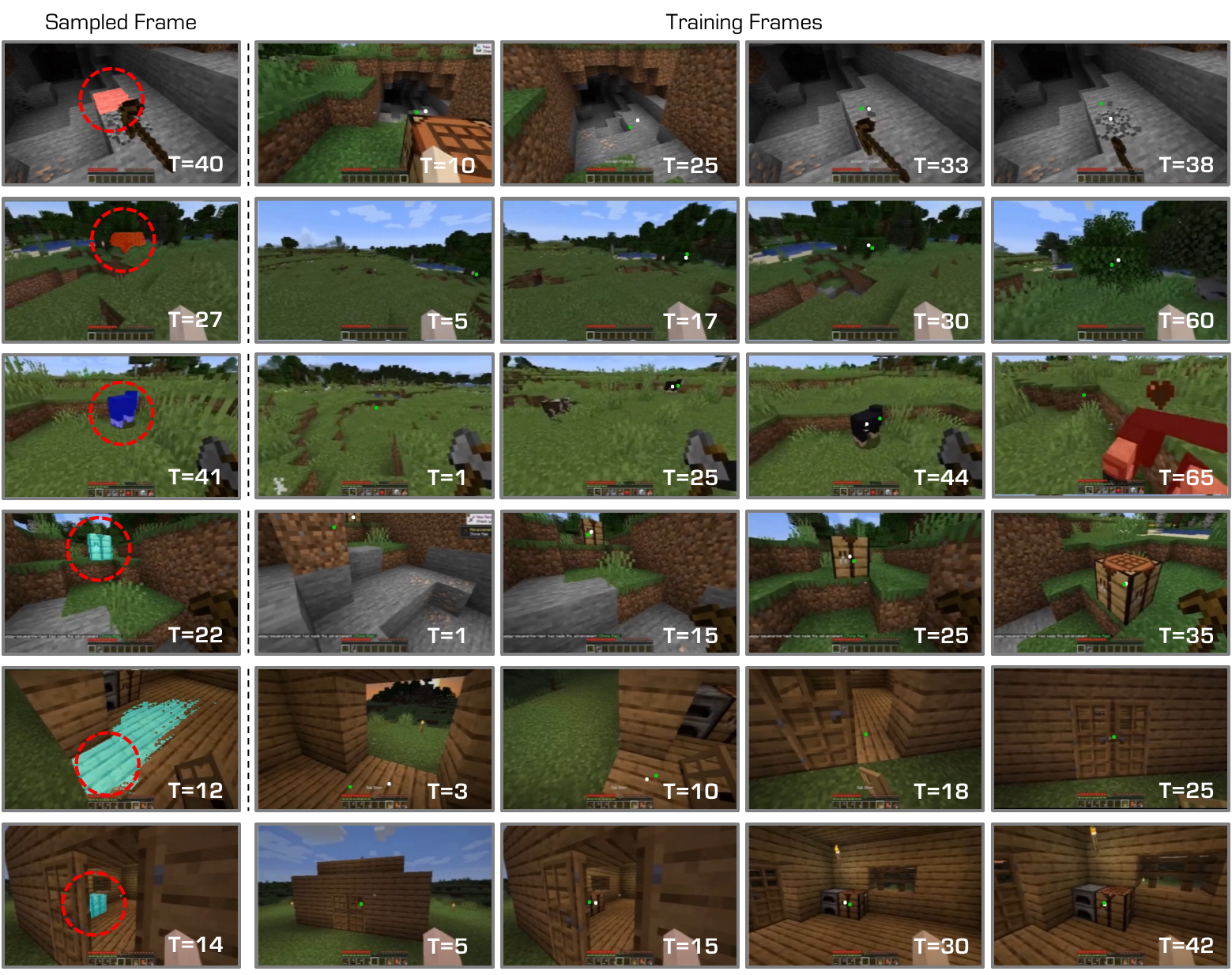}
\end{center}
\caption{
\textbf{Training Dataset.}
Each row represents a distinct interaction event (such as use, break, or approach) in which all segmentation masks correspond to the same goal. The left column presents sampled cross-view goals and their associated masks for each event, while the right column displays the corresponding frames during that event. This pairing of a cross-view objective with an egocentric action frame constitutes the foundation of our training dataset. In the images, the white point marks the center of the mask, whereas the green point indicates the location predicted by ROCKET-2.
}
\label{fig:dataset}
\end{figure*}

\end{document}